\documentclass[fleqn,10pt]{wlscirep}
\usepackage{booktabs} 
\usepackage[symbol]{footmisc}
\usepackage{multirow}
\usepackage{graphics}
\usepackage{graphicx}
\usepackage{subcaption}
\usepackage{tabularx}
\usepackage{color}
\usepackage{amsmath}
\usepackage{hyperref}
\usepackage{makecell}
\usepackage{caption}
\usepackage[ruled,vlined,linesnumbered]{algorithm2e}

\newcommand{\fullmodel}{\textbf{Guided Deep List}}
\newcommand{\skipnegative}{\textbf{SGNS}}
\newcommand{\skiph}{\textbf{SGHS}}
\newcommand{\gdlsgns}{\textbf{Guided Deep List (SGNS)}}
\newcommand{\gdlsghs}{\textbf{Guided Deep List (SGHS)}}
\newcommand{\baseline}{\textbf{Guidedlist}}

\title{Guided Deep List: Automating the Generation of Epidemiological Line Lists from Open Sources}
\author[1, *]{Saurav Ghosh}
\author[1]{Prithwish Chakraborty}
\author[2]{Bryan L. Lewis}
\author[3,4]{Maimuna S. Majumder}
\author[4]{Emily Cohn}
\author[4,5]{John S. Brownstein}
\author[2]{Madhav V. Marathe}
\author[1]{Naren Ramakrishnan}

\affil[1]{Discovery Analytics Center, Virginia Tech, Arlington, Virginia, USA,}
\affil[2]{Biocomplexity Institute, Virginia Tech, Arlington, Virginia, USA,}
\affil[3]{Massachusetts Institute of Technology, Cambridge, Massachusetts, USA,}
\affil[4]{Boston Children's Hospital, Boston, Massachusetts, USA,}
\affil[5]{Harvard Medical School, Boston, Massachusetts, USA.}
\affil[*]{sauravcsvt@vt.edu}
\date{\today}

\begin{document}

\begin{abstract}
Real-time monitoring and responses to emerging public health threats 
rely on the availability
of timely surveillance data. During the 
early stages of an epidemic, the ready availability of \textit{line lists} with 
detailed tabular information about laboratory-confirmed cases can assist  epidemiologists 
in making reliable inferences and forecasts. Such inferences are crucial to understand the epidemiology of a specific disease early enough to stop or control the outbreak.
However, construction of such line lists requires considerable human supervision and
therefore, difficult to generate in real-time. In this paper, we motivate {\fullmodel},
the first tool for building automated line lists (in near real-time) from open source reports of emerging
disease outbreaks. Specifically, we focus on deriving epidemiological characteristics of an 
emerging disease and the affected population from reports of illness. {\fullmodel}
uses distributed vector
representations (ala word2vec) to discover a set of indicators for each
line list feature. This discovery of indicators is followed by the use of
dependency parsing based techniques for final extraction in tabular form. We evaluate
the performance of {\fullmodel} against a human annotated line list provided by HealthMap 
corresponding to MERS outbreaks in Saudi Arabia. We demonstrate that {\fullmodel} 
extracts line list features with increased accuracy compared to a baseline method. We further
show how these automatically extracted line list features can be used for making
epidemiological inferences, such as inferring
demographics and symptoms-to-hospitalization period of affected individuals.
\end{abstract}

\maketitle

\section{Introduction}
\label{sec:intro}

An epidemiological line list~\cite{lau2014accuracy,majumder2014estimation} is a listing of individuals suffering from a disease 
that describes both their demographic details as well as the timing of clinically and epidemiologically significant events 
during the course of disease.  These are typically used during outbreak investigations of emerging diseases to identify key features, such as incubation period, 
symptoms, associated risk factors, and outcomes.  The ultimate goal is to understand the disease well enough to stop or control the outbreak. Ready availability 
of line lists can also be useful in contact tracing as well as risk identification of spread such as the spread of Middle Eastern Respiratory Syndrome (MERS) in 
Saudi Arabia or Ebola in West Africa.

Formats of line lists are generally dependent on the kind of disease being investigated. However, some  interesting features that are common for most formats 
include demographic information about cases. Demographic information can include age, gender, and location of infection. Depending on the disease being investigated, 
one can consider other addendums to this list, such as disease onset features (onset date, hospitalization date and outcome date) and clinical features 
(comorbidities, secondary contact, animal contact). 

Traditionally, line lists have been curated manually and have rarely been available to epidemiologists in near-real time. 
Our primary objective is to automatically generate line lists of emerging diseases from open source reports such as WHO bulletins~\cite{WHODONs} 
and make such lists readily available to epidemiologists. Previous work~\cite{lau2014accuracy,majumder2014estimation} 
has shown the utility in creating such lists through labor intensive human curation. We now seek to automate much of this effort. {\bf To the best of our knowledge,
our work is the first to automate the creation
of line lists.}

\begin{figure*}[!h]
  \centering
  \includegraphics[width=\linewidth]{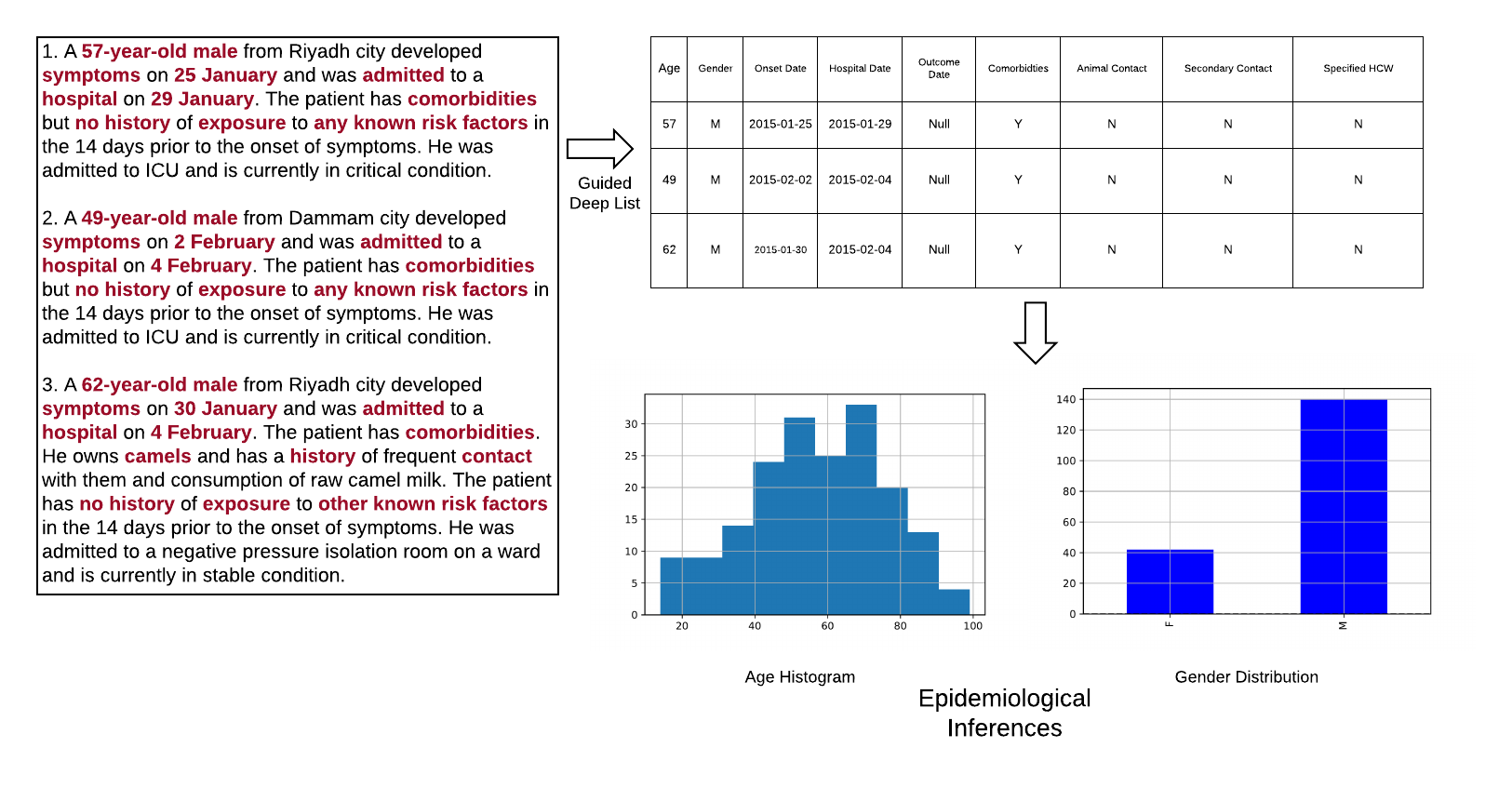}
  \vspace{-4em}
  \caption{Tabular extraction of line list by {\fullmodel} given a textual block of a WHO MERS bulletin. Each row in the extracted table depicts an infected case (or, patient) and columns represent the epidemiological features corresponding to each case. Information for each case in the table is then used to make epidemiological inferences, such as inferring demographic distribution of cases}
  \label{fig:ll_overall}
\end{figure*}

The availability of massive textual public health data
coincides with recent developments in text modeling, including
distributed vector representations such as
 word2vec~\cite{mikolovefficient,mikolovdistributed} and doc2vec~\cite{le2014distributed}. These neural network based language models when trained over a representative 
corpus convert words to dense low-dimensional vector representations, most popularly known as word embeddings. These word embeddings 
have been widely used with considerable accuracy to capture linguistic patterns and regularities, such as vec(\textit{Paris}) - vec(\textit{France}) $\approx$ vec(\textit{Madrid}) - vec(\textit{Spain})~\cite{mikolovregular,levyregular}. 
A second development relevant for line list generation
pertains to semantic dependency parsing, which has
 emerged as an effective tool
for information extraction, e.g., in
an open information extraction context~\cite{wu2010open}, 
Negation Detection~\cite{ou2015automatic,sohn2012dependency,ballesteros2012ucm}, relation extraction~\cite{bunescu2005shortest,levy2014dependency}
and event detection~\cite{muthiah2015planned}. Given an input sentence, dependency parsing is typically
used to extract its semantic tree 
representations where words are linked by directed edges called \textit{dependencies}. 

Building upon these techniques, we formulate {\fullmodel}, a novel framework for automatic 
extraction of line list from WHO bulletins~\cite{WHODONs}. 
{\fullmodel} is guided in the sense that the user provides a seed indicator (or, keyword) for each line list feature to guide
the extraction process. {\fullmodel} uses neural word embeddings to expand the seed indicator and 
generate a set of indicators for each line list feature. The set of indicators is subsequently provided as input 
to dependency parsing based shortest distance and negation detection approaches for extracting line list features. As can be seen in
Figure~\ref{fig:ll_overall}, {\fullmodel} takes a WHO bulletin as input and outputs epidemiological line list in tabular 
format where each row represents a line list case and each column depicts the features corresponding to each case.
The extracted line list provides valuable information to
model the epidemic and understand the segments of
population who would be affected.

Our main contributions are as follows.\\
\noindent $\>\bullet$ \textbf{Automated:} {\fullmodel} is fully automatic, requiring no human intervention.\\ 
\noindent $\>\bullet$ \textbf{Novelty:} To the best of our knowledge, there has been no prior systematic
efforts at tabulating such information automatically from publicly available health bulletins.\\
\noindent $\>\bullet$ \textbf{Real-time:} {\fullmodel} can be deployed for extracting line list in a (near) real-time setting.\\ 
\noindent $\>\bullet$ \textbf{Evaluation:} We present a detailed and prospective analysis of {\fullmodel} by evaluating 
  the automatically inferred line list against a human curated line list 
  for MERS outbreaks in Saudi Arabia. We also compare {\fullmodel} against 
  a baseline method.\\
\noindent $\>\bullet$ \textbf{Epidemiological inferences:} Finally, we also demonstrate some of the utilities of real-time automated line listing, such as inferring the demographic distribution and symptoms-to-hospitalization period.

\section{Problem Overview}
  \label{sec:prob}
  
  In this manuscript, we intend to focus on Middle Eastern Respiratory Syndrome (MERS) 
outbreaks in Saudi Arabia~\cite{majumder2014estimation} (2012-ongoing) as our case study. 
MERS was a relatively less understood disease when these outbreaks began. Therefore, MERS was poised as an emerging outbreak 
leading to good bulletin coverage about the infectious cases individually. This makes these disease outbreaks ideally suited to our goals. 
MERS is infectious as well and animal contact has been posited as one of the transmission mechanisms of the disease. For each line list case, we seek to extract automatically three types of epidemiological features as follows. (a) \textbf{Demographics:} Age and Gender, (b) \textbf{Disease onset:} onset date, hospitalization date and outcome date and (c) \textbf{Clinical features:} animal contact, secondary contact, comorbidities and specified healthcare worker (abbreviated as HCW).  

\begin{figure}[!h]
  \centering
  \includegraphics[width=\linewidth]{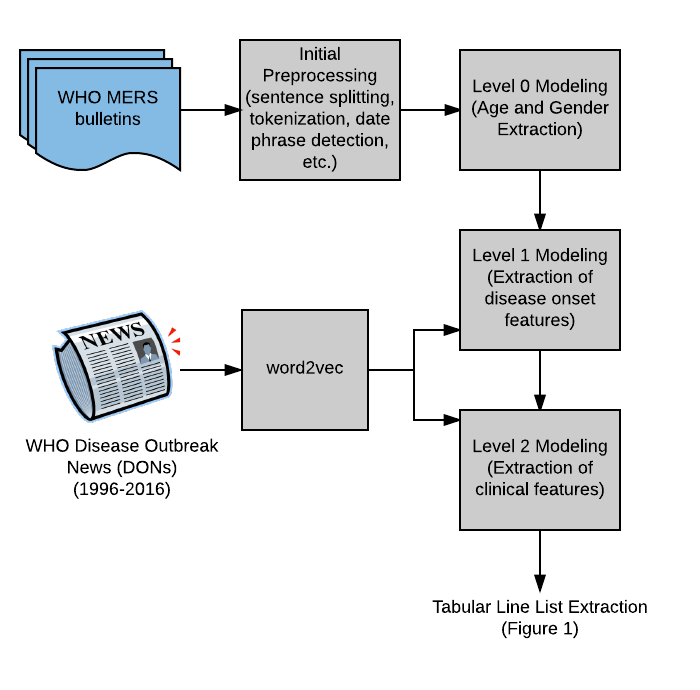}
  \vspace{-4em}
  \caption{Block diagram depicting all  components of the {\fullmodel} framework. Given multiple WHO MERS bulletins as input, 
  these components function in the depicted order to extract line lists in tabular form)}
  \label{fig:ll_block}
\end{figure}

In \figurename~\ref{fig:ll_block}, we show all the internal components comprising the 
framework of {\fullmodel}. {\fullmodel} takes multiple WHO MERS bulletins 
as input. The textual content of each bulletin is pre-processed by sentence splitting, tokenization, lemmatization, POS tagging, and date phrase detection using spaCy~\cite{spacycite} and BASIS Technologies’ Rosette Language Processing (RLP) tools~\cite{naren2014forecasting}. The pre-processing step is followed by three levels of modeling as follows. (a) Level 0 Modeling for extracting demographic information of cases, such as age and gender. In this level, we also identify the key sentences related to each line list case, (b) level 1 Modeling for extracting disease onset information and (c) level 2 Modeling for extracting clinical features. This is the final level of modeling in {\fullmodel} framework. Features extracted at this level are associated with two labels: \textit{Y} or \textit{N}. Therefore, 
modeling at this level combines neural word embeddings with dependency parsing-based negation detection approaches to classify the clinical features into \textit{Y} or \textit{N}. In the subsequent section, we will discuss each internal component of {\fullmodel} in detail. 

\section{Guided Deep List}
  \label{sec:methods}

Given multiple WHO MERS bulletins as input, {\fullmodel} proceeds through three levels 
of modeling for extracting line list features. We describe each level in turn.

\subsection{Level O Modeling}

In level 0 modeling, we extract the age and gender for each line list case.
These two features are mentioned in a reasonably structured way and therefore, can be 
extracted using a combination of regular expressions as shown in Algorithm~\ref{al:level0}. One of the primary challenges in extracting line list cases is the fact that a single WHO MERS bulletin can contain information about multiple cases. Therefore, there is a need to distinguish between cases mentioned in the bulletin. In level 0 modeling, we make use of the age and gender extraction to also identify sentences associated with each case. Since age and gender are the fundamental information to be recorded for a line 
list case, we postulate that the sentence mentioning the age and gender 
will be the starting sentence describing a line list case (see the textual block in Figure~\ref{fig:ll_overall}). Therefore, the number of cases mentioned in the bulletin will be equivalent to the number of sentences mentioning age and gender information. We further postulate that information related to the other features (disease onset or critical) will be present either in the starting sentence or the sentences subsequent to the starting one not mentioning any age and gender related information ((see the textual block in Figure~\ref{fig:ll_overall})). For more details
on level 0 modeling, please see Algorithm~\ref{al:level0}. In Algorithm~\ref{al:level0}, $\mathcal{N}$ represents the number of line list cases mentioned in the bulletin and $\mathcal{SC}_{n}$ represents the
set of sentences mentioning the $n^{th}$ case.
  
\begin{algorithm}
   \DontPrintSemicolon
   \SetKwInOut{Input}{Input}\SetKwInOut{Output}{Output}
    \Input{set of sentences in the input WHO MERS bulletin}
    \Output{Age and Gender for each line list case, index of the starting sentence for each case}
    n = 0;\\
    $\mathcal{SC}_{n}$ = Null;\\
    $\mathcal{R}_{1}$ = \verb@\s+(?P<age>\d{1,2})(.{0,20})(\s+|-)(?P<gender>woman|man|male|female|boy|girl|housewife)@;\\

    $\mathcal{R}_{2}$ = \verb@\s+(?P<age>\d{1,2})\s*years?(\s|-)old@;\\

    $\mathcal{R}_{3}$ = \verb@\s*(?P<gender>woman|man|male|female|boy|girl|housewife|he|she)@;\\
    \For{each sentence in the bulletin}{
        is-starting $\rightarrow$ 0;\\
        \If{$\mathcal{R}_{1}$.match(sentence)}{
             Age = int($\mathcal{R}_{1}$.groupdict()['age']);\\
             Gender = $\mathcal{R}_{1}$.groupdict()['gender'];\\
             is-starting $\rightarrow$ 1;\\
        }
        \Else{
            \If{$\mathcal{R}_{2}$.match(sentence)}{
                Age = int($\mathcal{R}_{3}$.groupdict()['age']);\\
             }
             \Else{
                 Age = Null;\\
             }
            \If{$\mathcal{R}_{3}$.match(sentence)}{
                Gender = int($\mathcal{R}_{3}$.groupdict()['gender']);\\
            }
            \Else{
                Gender = Null;\\
            }
            \If{Age $\neq$ Null $\&\&$ Gender $\neq$ Null}{
                is-starting $\rightarrow$ 1;\\
            }
         }
         \If{is-starting}{
             n += 1;\\
             $\mathcal{SC}_{n}$ = index of the sentence;\\
          }
    }
    $\mathcal{N}$ = n;
    \caption{Level 0 modeling\label{al:level0}}
\end{algorithm}

\subsection{WHO Template Learning}
Before presenting the details of level 1 modeling and level 2 modeling, we will 
briefly discuss the WHO template learning process which provides word
embeddings as input to both these levels of modeling (see \figurename~\ref{fig:ll_block}). 
In the template learning process, our main objective is to identify words which 
tend to share similar contexts or appear in the contexts of each 
other specific to the WHO bulletins (contexts of a word refer to the words surrounding 
it in a specified window size). For instance, consider the sentences $\mathcal{S}_{1} = $\textbf{\textit{The patient had no contact with animals}} and $\mathcal{S}_{2} =$\textbf{\textit{The patient was supposed to have no contact with camels}}. The terms \textit{animals} and \textit{camels} appear in similar contexts in both $\mathcal{S}_{1}$ and $\mathcal{S}_{2}$. Both the terms \textit{animals} and \textit{camels} are indicative of information pertaining to patient's exposure to animals or animal products.

Similarly, consider the sentences $\mathcal{S}_{3} = $\textbf{\textit{The patient had an onset of symptoms on 23rd January 2016}} and $\mathcal{S}_{4} = $\textbf{\textit{The patient developed symptoms on 23rd January 2016}}. The terms \textit{onset} and \textit{symptoms} are indicators for the onset date feature and both of them appear in similar contexts or contexts of each other in $\mathcal{S}_{3}$ and $\mathcal{S}_{4}$. 

For the template learning process, neural network inspired word2vec models are ideally suited to our goals because these models work on the hypothesis that words sharing similar contexts or tending 
to appear in the contexts of each other have similar embeddings. In recent years, word2vec models based on the skip-gram 
architectures~\cite{mikolovefficient,mikolovdistributed} have emerged as the most popular word embedding models for information extraction tasks~\cite{levyparam,levydependency,ghoshcikm}.
We used two variants of skip-gram models: (a) the skip-gram model trained using the negative sampling technique 
({\skipnegative}~\cite{mikolovdistributed}) and (b) the skip-gram model trained using hierarchical sampling ({\skiph}~\cite{mikolovdistributed}) 
to generate embeddings for each term in the WHO vocabulary $\mathcal{W}$. $\mathcal{W}$ refers to the list of all unique terms extracted 
from the entire corpus of WHO Disease Outbreak News (DONs) corresponding to all diseases downloaded from \url{http://www.who.int/csr/don/archive/disease/en/}. 
The embeddings for each term in $\mathcal{W}$ were provided as input to level 1 modeling and level 2 modeling as shown in \figurename~\ref{fig:ll_block}.
 
\subsection{Level 1 Modeling}
\label{sec:level1}

The level 1 modeling is responsible for extracting the disease onset features, such as symptom onset date, hospitalization date 
and outcome date for each linelist case, say the $n^{th}$ case. For extracting a given 
disease onset feature, the level 1 modeling takes three inputs: (a) seed indicator for the feature, (b) 
the word embeddings generated using {\skipnegative} or {\skiph} for each term in the WHO vocabulary $\mathcal{W}$ 
and (c) $\mathcal{SC}_{n}$ representing the set of sentences describing the $n^{th}$ case for which we are extracting
the feature.

\noindent \paragraph*{\textbf{Growth of seed indicator}} In the first phase of level 1 modeling, 
  we discover the top-$K$ similar (or, closest) indicators in the embedding space 
  to the seed indicator for each feature. The similarity metric used is the standard
  cosine similarity metric. Therefore, we expand the seed indicator to create a set of $K+1$ 
  indicators for each feature. In Table~\ref{fig:seed} we show the indicators discovered by {\skipnegative} 
  for each disease onset feature given the seed indicators as input. 

\begin{table}[!ht]
\centering
\caption{Seed indicator and the discovered indicators using word embeddings generated by {\skipnegative}}
\label{fig:seed}
\begin{tabular}{|c|c|c|}
  \hline
  Features & Seed indicator & Discovered indicators \\ \hline
Onset date & onset & \begin{tabular}[c]{@{}c@{}}symptoms, symptom, prior,\\days, dates\end{tabular} \\ \hline
Hospitalization date & hospitalized & \begin{tabular}[c]{@{}c@{}}admitted, screened, hospitalised,\\ passed, discharged\end{tabular} \\ \hline
Outcome date & died & \begin{tabular}[c]{@{}c@{}}recovered, passed, became,\\ ill, hospitalized\end{tabular} \\ \hline
\end{tabular}
\end{table}

\begin{figure}
\centering
\includegraphics[width=\linewidth,scale=0.05]{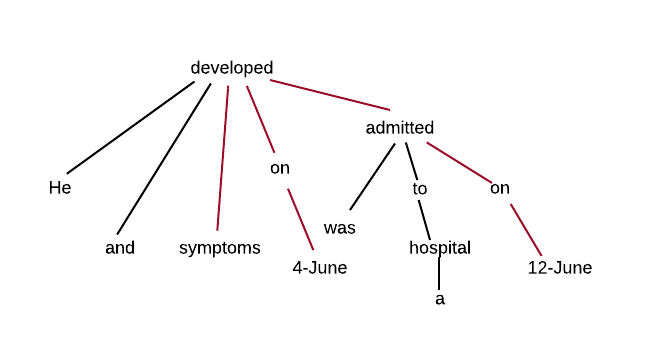}
\vspace{-2em}
\caption{Undirected dependency graph corresponding to $\mathcal{S}_{5}$. The red-colored edges depict those 
edges included in the shortest paths between the date phrases (\textit{4-June}, \textit{12-June}) and the
indicators (\textit{symptoms}, \textit{admitted})}
\label{fig:undep}
\end{figure}

\noindent \paragraph*{\textbf{Shortest Dependency Distance}} In the second phase, we use these $K+1$ indicators to extract the 
  disease onset features. For each indicator $\mathcal{I}_{t} \forall t \in 1,2,\ldots,K+1$, we identify the sentences mentioning
  $\mathcal{I}_{t}$ by iterating over each sentence in $\mathcal{SC}_{n}$. Then, for each sentence mentioning $\mathcal{I}_{t}$, 
  we discover the shortest path along the undirected dependency graph between $\mathcal{I}_{t}$ and the date phrases mentioned in 
  the sentence. Subsequently, we calculate the length of the shortest path as the number of edges encountered while traversing along 
  the shortest path. The length of the shortest path is referred to as the \textit{dependency distance}. E.g., consider the sentence 
  $\mathcal{S}_{5} = $ \textbf{\textit{He developed symptoms on 4-June and was admitted to a hospital on 12-June.}} The sentence $\mathcal{S}_{5}$ containes the date phrases \textit{4-June} and \textit{12-June}. $\mathcal{S}_{5}$
  also contains the indicator \textit{symptoms} for onset date and \textit{admitted} for hospitalization date (see Tables~\ref{fig:seed}).
  In Figure~\ref{fig:undep}, we show the undirected dependency graph for $\mathcal{S}_{5}$. We observe that the \textit{dependency distance}
  from \textit{symptoms} to \textit{4-June} is 3 (\textit{symptoms} $\rightarrow$ \textit{developed} $\rightarrow$ \textit{on} $\rightarrow$ \textit{4-June})
  and \textit{12-June} is 4 (\textit{symptoms} $\rightarrow$ \textit{developed} $\rightarrow$ \textit{admitted} $\rightarrow$ \textit{on} $\rightarrow$ \textit{12-June}).
  Similarly, the dependency distance from \textit{admitted} to \textit{4-June} is 3 (\textit{admitted} $\rightarrow$ \textit{developed} $\rightarrow$ \textit{on} $\rightarrow$ \textit{4-June}) 
  and \textit{12-June} is 2 (\textit{admitted} $\rightarrow$ \textit{on} $\rightarrow$ \textit{4-June}).
  Therefore, for each indicator we extract a set of date phrases and the dependency distance corresponding to each date phrase. 
  The output value of the indicator is set to be the date phrase located at the shortest dependency distance. E.g., in
  $\mathcal{S}_{5}$, the output values of \textit{symptoms} and \textit{admitted} will be \textit{4-June} and 
  \textit{12-June} respectively. The final output for each disease feature is obtained by performing majority voting 
  on the outputs of the indicators. For more algorithmic details, please see Algorithm~\ref{al:level1}.

\begin{algorithm}
    \DontPrintSemicolon
    \SetKwInOut{Input}{Input}\SetKwInOut{Output}{Output}
    \Input{seed indicator, word embeddings for each term in $\mathcal{W}$, $\mathcal{SC}_{n}$}
    \Output{date phrase}
    Growth of seed indicator using word embeddings to generate $K+1$ indicators represented as $\mathcal{I}_{t} \forall t \in 1,2,\ldots,K+1$;\\
    \For{each $\mathcal{I}_{t}$}{
      dependency-dist = dict(); empty dictionary\\
      \For{each sentence in $\mathcal{SC}_{n}$}{ 
        check the mention of $\mathcal{I}_{t}$;\\
        \If{$\mathcal{I}_{t}$ found}{
            Identify the date phrases mentioned in the sentence;\\
            \If{at least one date phrase is found}{
                 construct the undirected dependency graph for the sentence (see Figure~\ref{fig:undep});\\
                 \For{each date phrase in the sentence}{
                    dependency-dist[date phrase] = dependency distance (see section~\ref{sec:level1});\\
                 }
            }
            \Else{
              continue;\\
            }
                    
       }
       \Else{
         continue;\\
       }
    }
    Output of $\mathcal{I}_{t} = $ date phrase in dependency-dist having the shortest dependency distance;\\
    }
    final output = majority voting on the outputs of each $\mathcal{I}_{t}$;\\
\caption{Level 1 modeling\label{al:level1}}
\end{algorithm}

\subsection{Level 2 Modeling}
\label{sec:level2}

The level 2 modeling is responsible for extracting the clinical features for each line list case. 
Extraction of clinical features is a binary classification problem where we have to classify each feature into two classes - \textit{Y} 
or \textit{N}. The first phase of level 2 modeling is similar to level 1 modeling. Seed indicator for each clinical feature is provided as 
input to the level 2 modeling and we extract the $K+1$ indicators for each such feature by discovering the top-$K$ most similar indicators 
to the seed indicator (in terms of cosine similarities) using the word embeddings generated during the WHO template learning process.   

\begin{figure}
\centering
\includegraphics[width=\linewidth,scale=0.05]{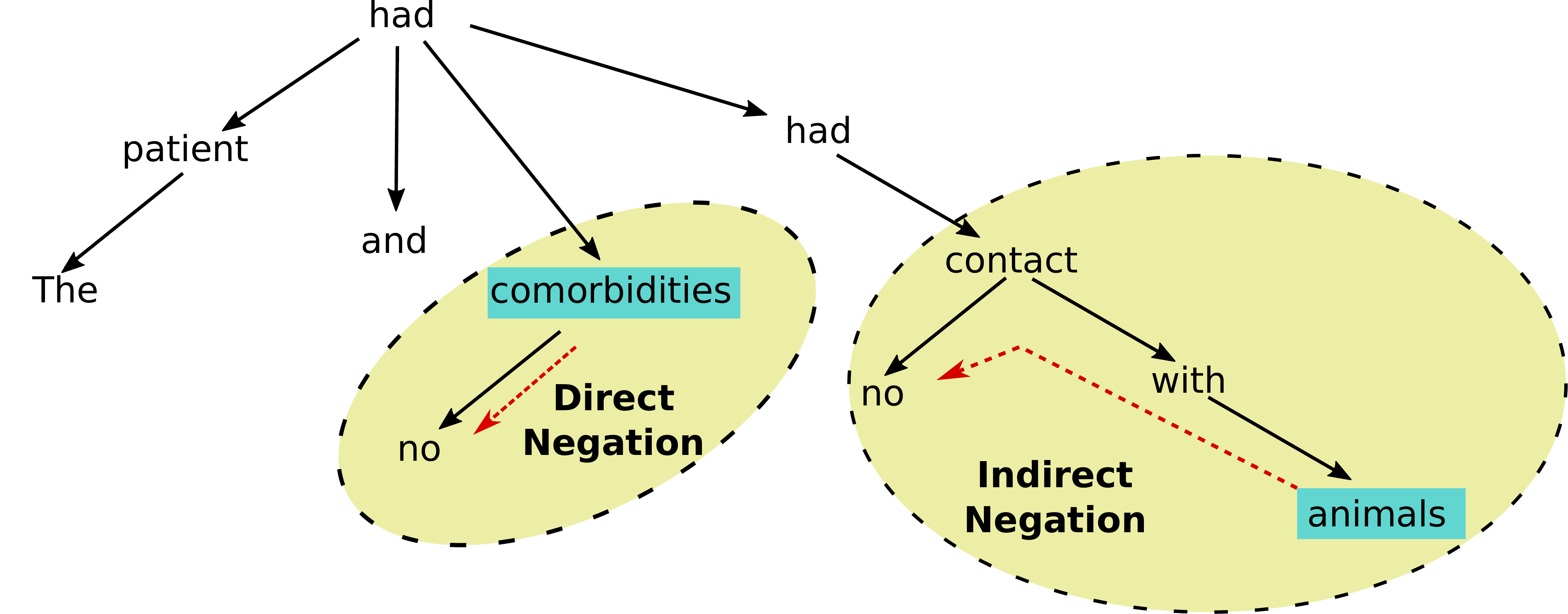}
\caption{Directed dependency graph corresponding to $\mathcal{S}_{6}$ showing direct and indirect negation detection}
\label{fig:negdep}
\end{figure}

\noindent \paragraph*{\textbf{Dependency based negation detection}} In the second phase, we 
make use of the $K+1$ indicators extracted in the first phase and a static lexicon of 
negation cues~\cite{diaz2012ucm}, such as \textit{no}, \textit{not}, \textit{without}, 
\textit{unable}, \textit{never}, etc.\ to detect negation for a clinical feature. If no
negation is detected, we classify the feature as \textit{Y}, otherwise \textit{N}. 
For each indicator $\mathcal{I}_{t} \forall t \in 1,2,\ldots,K+1$, we identify the 
first sentence (referred to as $\mathcal{S}_{\mathcal{I}_{t}}$) mentioning $\mathcal{I}_{t}$
by iterating over the sentences in $\mathcal{SC}_{n}$. Once $\mathcal{S}_{\mathcal{I}_{t}}$ 
is identified, we perform two types of negation detection on the directed dependency 
graph $\mathcal{D}_{\mathcal{I}_{t}}$ constructed for $\mathcal{S}_{\mathcal{I}_{t}}$.
\\
\textbf{Direct Negation Detection:} In this negation detection, we search for a negation cue among the 
neighbors of $\mathcal{I}_{t}$ in $\mathcal{D}_{\mathcal{I}_{t}}$. If a negation cue is found, then the output of 
$\mathcal{I}_{t}$ is classified as \textit{N}.
\\
\textbf{Indirect Negation Detection.} Absence of a negation cue in the neighborhood of $\mathcal{I}_{t}$
drives us to perform indirect negation detection. In this detection, we locate those terms in 
$\mathcal{D}_{\mathcal{I}_{t}}$ for which $\mathcal{D}_{\mathcal{I}_{t}}$ has a directed path from each of 
these terms as source to $\mathcal{I}_{t}$ as target. We refer to these terms as the predecessors of $\mathcal{I}_{t}$ 
in $\mathcal{D}_{\mathcal{I}_{t}}$. Then, we search for negation cues in the neighborhood of each predecessor. If we 
find a negation cue around a predecessor, we assume that the indicator $\mathcal{I}_{t}$ is also affected
by this negation and we classify the output of $\mathcal{I}_{t}$ as \textit{N}. For example, consider the sentence $\mathcal{S}_{6} = $\textbf{\textit{The patient had no comorbidities and had no contact with animals.}} and the directed dependency graph corresponding to 
$\mathcal{S}_{6}$ is shown in Figure~\ref{fig:negdep}. Sentence $\mathcal{S}_{6}$ contains the seed indicators $\textit{comorbidities}$ for comorbidities and 
$\textit{animals}$ for animal contact. In Figure~\ref{fig:negdep}, 
we observe direct negation detection for comorbidities as the negation cue \textit{no} is located in the 
neighborhood of the indicator \textit{comorbidities}. However, for animal contact, we observe indirect negation
detection as the negation cue \textit{no} is situated in the neighborhood of the term \textit{contact} which is one of the 
predecessors of the indicator \textit{animals}.
 
Therefore, for a clinical feature we have $K+1$ indicators
and the classification output \textit{Y} or \textit{N} from each indicator. The final 
output for a feature is obtained via majority voting on the outputs of the indicators.

\begin{algorithm}
    \DontPrintSemicolon
    \SetKwInOut{Input}{Input}\SetKwInOut{Output}{Output}
    \Input{seed indicator, word embeddings for each term in $\mathcal{W}$, negation cues, $\mathcal{SC}_{n}$}
    \Output{\textit{Y} or \textit{N}}
    Growth of seed indicator using word embeddings to generate $K+1$ indicators represented as $\mathcal{I}_{t} \forall t \in 1,2,\ldots,K+1$;\\
    \For{each $\mathcal{I}_{t}$}{
    Iterate over each sentence in $\mathcal{SC}_{n}$ and identify the first sentence $\mathcal{S}_{{I}_{t}}$ mentioning ${I}_{t}$;\\
    Construct the directed dependency graph $\mathcal{D}_{{I}_{t}}$ (see Figure~\ref{fig:negdep}) for $\mathcal{S}_{{I}_{t}}$;\\
    $N_{{I}_{t}} = $ set of terms connected to $\mathcal{I}_{t}$ in $\mathcal{D}_{{I}_{t}}$, i.e.\ neighbors of $\mathcal{I}_{t}$;\\
    $P_{{I}_{t}} = $ predecessors of $\mathcal{I}_{t}$ in $\mathcal{D}_{{I}_{t}}$;\\
    Isnegation $\leftarrow 0$;\\
    \If{$N_{{I}_{t}}$ has a negation cue}{
         output of $\mathcal{I}_{t}$= \textit{N};\\   
         Isnegation $\leftarrow 1$;\\
         break;
         }
      \Else{
            Iterate over each term in $P_{{I}_{t}}$ and seach for a negation cue in the neighborhood;\\
            \If{negation cue found in neighborhood of a predecessor}{
                output of $\mathcal{I}_{t}$= \textit{N};\\
                Isnegation $\leftarrow 1$;\\
                break;\\
                }
            }
       \If{$\neg$Isnegation}{
              output of $\mathcal{I}_{t}$= \textit{Y};\\  
       }
    }
    final output = majority voting on the outputs of each $\mathcal{I}_{t}$;\\
    \caption{Level 2 modeling\label{al:level2}}
\end{algorithm}

\section{Experimental Evaluation}
  \label{sec:expts}

In this section, we first provide a brief description of our experimental
setup, including the models for automatic extraction of line lists, human 
annotated line lists, accuracy metric and parameter settings.

\subsection{WHO corpus}

The WHO corpus used in the template learning process (see \figurename~\ref{fig:ll_block})
was downloaded from \url{http://www.who.int/csr/don/archive/disease/en/}. The corpus contains outbreak news articles related to a wide range of diseases reported during the time period 1996 to 2016. The textual content of each article was pre-processed by sentence splitting, tokenization and lemmatization using spaCy~\cite{spacycite}. After pre-processing, the WHO corpus was found to contain 35,485 sentences resulting in a vocabulary $\mathcal{W}$ of 4447 words. 

\subsection{Models}

We evaluated the following automated line listing models.

\noindent $\>\bullet$ {\gdlsgns}: Variant of {\fullmodel} with {\skipnegative} used as the word2vec model in the WHO template learning process.\\
\noindent $\>\bullet$ {\gdlsghs}: Variant of {\fullmodel} with {\skiph} used as the word2vec model in the WHO template learning process.\\
\noindent {\baseline}: Baseline model which does not use any word embedding
model (absence of WHO template learning) to expand the seed indicator in order to generate $K+1$ indicators for each feature. Therefore, {\baseline} uses only a single indicator (seed indicator) to extract line list features.\\

\subsection{Human annotated line list}

\begin{figure}[t!]
\centering
\includegraphics[width=\linewidth]{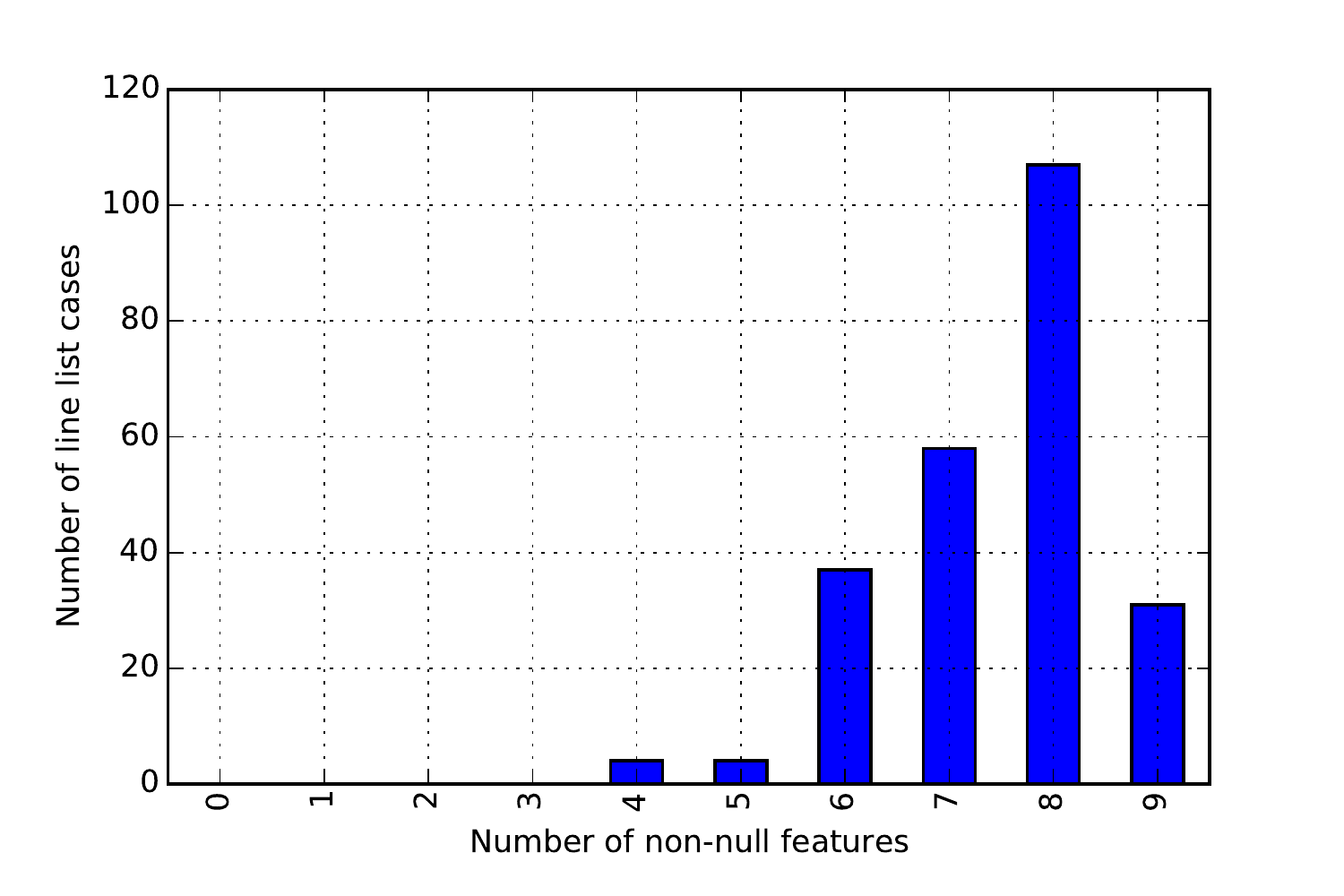}
\caption{Distribution of non-null features in the human annotated line list}
\label{fig:nonnull}
\end{figure}

We evaluated the line list extracted by the automated line listing models
against a human annotated line list for MERS outbreaks in Saudi Arabia.
To create the human annotated list, patient and outcome data for confirmed MERS cases were collected from the MERS Disease Outbreak News (DONs) reports of WHO~\cite{WHODONs} and curated into 
a machine-readable tabular line list. In the human annotated list, 
total number of confirmed cases were 241 curated from 64 WHO bulletins reported during the period October 2012 to February 2015. Some of these 241 cases have missing (null) features (see Figure~\ref{fig:ll_overall}). In Figure~\ref{fig:nonnull}, we show the distribution of non-null 
features in the human annotated list. We observe that majority of human annotated cases have at least 6 (out of 
9) non-null features with the peak of the distribution at 8.   

\subsection{Accuracy metric}

\paragraph*{\textbf{Matching automated line list to human annotated list.}} For evaluation, the problem is: we are given a set of automated line list cases and a set of human annotated cases for a single WHO MERS bulletin. Our strategy is to costruct a bipartite graph~\cite{naren2014forecasting} where (i) an edge exists if the automated case and the human annotated case is extracted from the same 
WHO bulletin and (ii) the weight on the edge denotes the quality score (QS). Quality score (QS) is defined as the number of correctly extracted features in the automated case divided by the number of non-null features in the human annotated case. We then construct a maximum weighted bipartite matching~\cite{naren2014forecasting}. Such matchings are conducted for each WHO bulletin to extract a set of matches where each match represents a pair (automated case, human annotated case) and is also associated with a QS\@. Once the matches are found for all the WHO bulletins, we computed the average QS by averaging the QS values across the matches.

Once the average QS and QS for each match are computed, we also computed the accuracy for each line list feature.  
For the demographic and disease onset features, we computed the accuracy classification score using 
scikit-learn~\cite{scikit-learn} by comparing the automated features against the human annotated features across 
the matches. The clinical features are associated with two classes - \textit{Y} and \textit{N} (see Figure~\ref{fig:ll_overall}). 
For each class, we computed the F1-score using scikit-learn~\cite{scikit-learn} where F1-score can be interpreted as a 
harmonic mean of the precision and recall. F1-score reaches its best value at 1 and worst score at 0. Along with the F1-score
for each class, we also report the average F1-score across the two classes.

\subsection{Parameter settings}

Each variant of {\fullmodel} inherits the parameters of the word embedding models as shown in Table~\ref{tab:ben_param}.  
Apart from the word embedding parameters, {\fullmodel} also inherits the parameter $K$ which 
refers to the $K+1$ indicators for disease onset or clinical features (see Section~\ref{sec:methods}).
In Table~\ref{tab:ben_param}, we provide the list of all parameters, the explored values for
each parameter and the applicable models corresponding to each parameter. We selected the optimal parameter configuration 
for each model based on the maximum average QS value as well as maximum average of the individual feature accuracies across the matches.

\section{Results}
  \label{sec:results}

In this section we try to ascertain the efficacy and applicability of
{\fullmodel} by investigating some of the pertinent questions related to
the problem of automated line listing.

\par{\textbf{Multiple indicators vs single indicator - which is the better method for automated line listing?}}

\begin{figure}[t!]
\centering
\includegraphics[width=\linewidth]{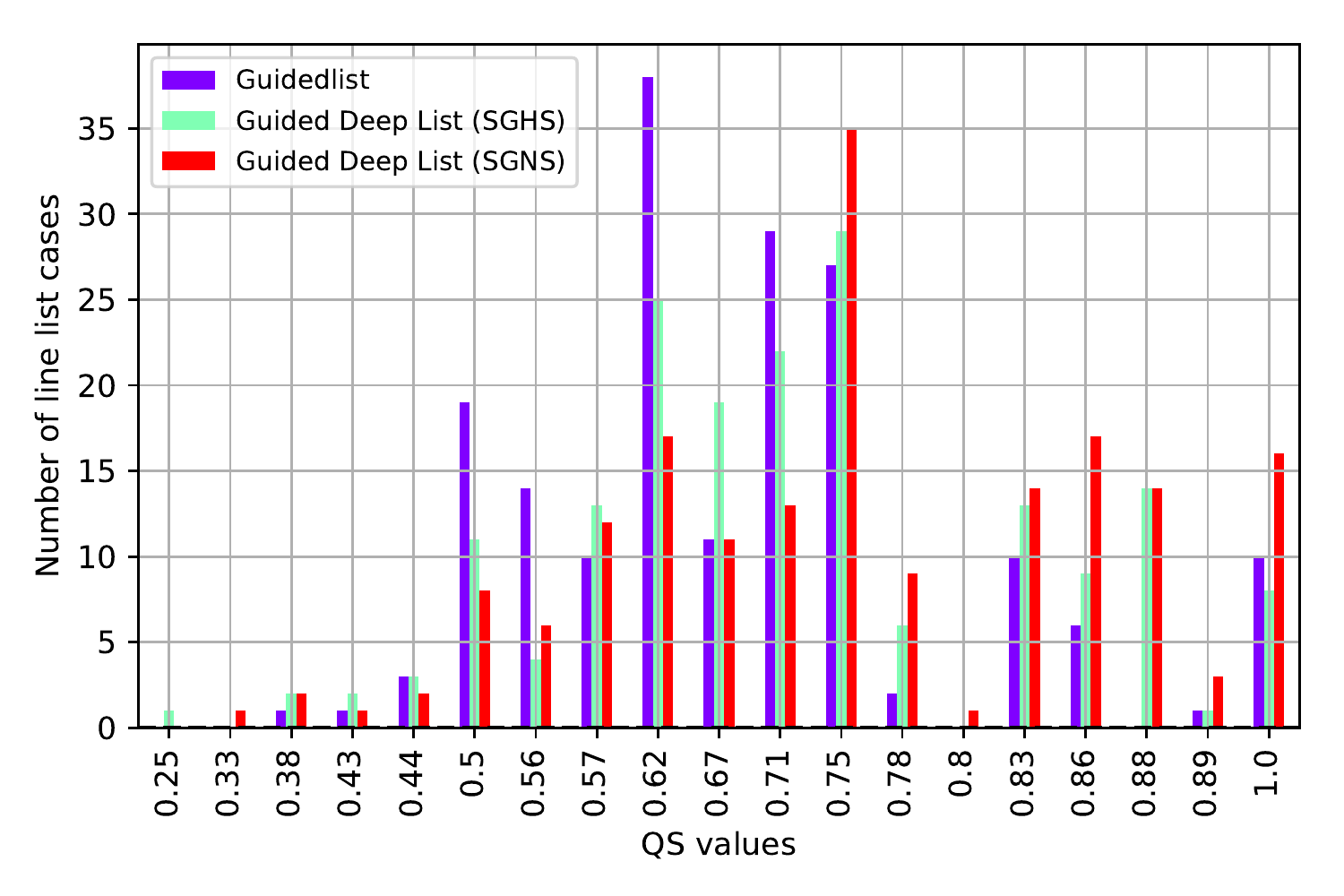}
\vspace{-3em}
\caption{Distribution of QS values for each automated line listing model corresponding 
to MERS line list in Saudi Arabia. X-axis represents QS values and Y-axis represents the number of
automated line list cases having a particular QS value}
\label{fig:coverage}
\end{figure}

As mentioned in section~\ref{sec:expts}, {\gdlsgns} and {\gdlsghs} uses multiple indicators discovered by
word2vec, whereas the baseline {\baseline} uses only the seed indicator to infer line list features. 
We executed our automated line listing models taking as input the same set of 64 WHO MERS bulletins from which 
241 human annotated line list cases were extracted. In Table~\ref{tab:qs}, we observe that the number of automated line list cases (198) and the matches (182) after maximum bipartite matching is same for all the models. This is due to the reason that level 0 modeling (age and gender extraction) is the common modeling component in all the models and the number of extracted line list cases depends on the age and gender extraction (see section~\ref{sec:methods}). In Table~\ref{tab:qs}, we also compared the average QS achieved by each model. We observe that
{\gdlsgns} is the best performing model achieving an average QS of 0.74 over {\gdlsghs} (0.71) and {\baseline} (0.67).
To further validate the results in Table~\ref{tab:qs}, we also show the QS distribution for each model in Figure~\ref{fig:coverage}
where x-axis represents the QS values and the y-axis represents the number of automated line list cases having a particular QS
value. For {\baseline}, the peak of QS distribution is at 0.62. However, for {\gdlsgns} and {\gdlsghs}, the peak of the distribution is at 0.75.
We further observe that {\gdlsgns} extracts higher number of line list cases with a perfect QS of 1 in comparison to {\baseline}. 

We also compared the models on the basis of individual accuracies of the line list features across the matches 
in Tables~\ref{tab:acc_table} and~\ref{tab:f1}. In Table~\ref{tab:acc_table}, all the models achieve similar performance for
the demographic features since level 0 modeling is similar
for all the models (see section~\ref{sec:methods}). However, for the disease onset features, both {\gdlsgns} and {\gdlsghs} outperform the baseline achieving an average accuracy of $0.45$ and $0.43$ in comparison to {\baseline} ($0.12$) respectively. {\gdlsgns} is the best performing model for onset date. However, for hospitalization date and outcome date,
{\gdlsghs} is the better performing model than {\gdlsgns}. In Table~\ref{tab:f1}, for the clinical features, we observe that {\gdlsgns} performs better than {\gdlsghs} and {\baseline} for comorbidities and specified HCW on the basis of average F1-score. 
Specifically, for specified HCW, {\gdlsgns} outperforms {\gdlsghs} and {\baseline} for the minority class \textit{Y}. For animal
contact, {\gdlsghs} emerges out to be the best performing model in terms of average F1-score, specifically outperforming the
competing models for the minority class \textit{Y}. {\baseline} only performs better for secondary contact, even though
the performance for the minority class \textit{Y} is almost similar to {\gdlsghs} and {\gdlsgns}. Overall, we can conclude 
from Table~\ref{tab:f1} that {\fullmodel} employing multiple indicators discovered via {\skipnegative} or {\skiph}
shows superior performance than {\baseline} in majority of the scenarios, specifically for the minority class of each clinical feature. 
To further validate the results in Table~\ref{tab:f1}, the confusion matrix for each model and each clinical feature can be found in \url{https://github.com/sauravcsvt/KDD_linelisting}.

\begin{table}[!ht]
  \small
\centering
\caption{Average Quality Score (QS) achieved by each automated line listing model 
for MERS line list in Saudi Arabia. As can be seen, {\gdlsgns} shows best performance 
achieving an average QS of 0.73}
\label{tab:qs}
\begin{tabular}{|c|c|c|c|c|}
\hline
Models & Human lists & Auto lists & Matches & Average QS \\ \hline
\makecell{{\baseline} \\ (baseline)} & 241 & 198 & 182 & 0.67 \\ \cline{1-5} 
\makecell{{\fullmodel} \\ ({\skiph})} & 241 & 198 & 182 & 0.71 \\ \cline{1-5} 
\makecell{{\fullmodel} \\ (\skipnegative)}  & 241 & 198 & 182 & \textbf{0.74} \\ \hline
\end{tabular}
\end{table}

\begin{table}[!ht]
\centering
\caption{Comparing the automated line listing models based on the accuracy
score for the demographics and disease onset features. For the
disease onset features, {\gdlsgns} emerges out to be the best performing model. 
However, for the demographic features, all the models achieve almost similar 
performance}
\label{tab:acc_table}
\begin{tabular}{|c|c|c|c|c|}
  \hline
  \makecell{Feature \\ type} & Features & \makecell{{\baseline}\\ (baseline)} & \makecell{{\fullmodel} \\ ({\skiph})} & \makecell{{\fullmodel} \\ ({\skipnegative})} \\ \hline
\multirow{2}{*}{Demographics} & Age & 0.87 & \textbf{0.91} & 0.87 \\ \cline{2-5}
  & Gender & \textbf{0.99} & 0.98 & 0.97 \\ \cline{2-5}
  & Average & 0.93 & \textbf{0.95} & 0.92 \\ \cline{1-5}
  \multirow{3}{*}{\makecell{Disease \\ onset}} & Onset date & 0.01 & 0.01 & \textbf{0.37} \\ \cline{2-5}
  & Hospitalization date & 0.11  & \textbf{0.63} & 0.62 \\ \cline{2-5}
  & Outcome date & 0.48 & \textbf{0.66} & 0.36 \\ \cline{2-5}
  & Average & 0.20 & 0.43 & \textbf{0.45} \\ \hline
\end{tabular}
\end{table}

\begin{table}[!ht]
\centering
\caption{Comparing the performance of the automated line listing models for extracting 
clinical features corresponding to MERS line list in Saudi Arabia. We report the 
F1-score for class Y, class N and average F1-score across the two classes. For animal contact,
{\gdlsghs} emerges out to be the best performing model. For comorbidities and specified HCW,
{\gdlsgns} shows best performance. However, for secondary contact, {\baseline} achieve superior 
performance in comparison to {\fullmodel}}
\label{tab:f1}
\begin{tabular}{|c|c|c|c|c|}
\hline
\makecell{Clinical Feature \\ (Y:N)} & Class & \makecell{{\baseline} \\ (baseline)} & \begin{tabular}[c]{@{}c@{}}{\fullmodel}\\ ({\skiph})\end{tabular} & \begin{tabular}[c]{@{}c@{}}{\fullmodel}\\ ({\skipnegative})\end{tabular} \\ \hline
\multirow{3}{*}{\makecell{Animal contact \\ (1:3)}} & Y & 0.33 & \textbf{0.68} & 0.37 \\ \cline{2-5} 
  & N & 0.87 & \textbf{0.91} & 0.88 \\ \cline{2-5} 
  & Average & 0.60 & \textbf{0.79} & 0.63 \\ \cline{1-5} 
  \multirow{3}{*}{\makecell{Secondary contact \\ (1:3)}} & Y & \textbf{0.57} & 0.52 & 0.56 \\ \cline{2-5} 
  & N & \textbf{0.86} & 0.70 & 0.72 \\ \cline{2-5} 
  & Average & \textbf{0.71} & 0.61 & 0.64 \\ \cline{1-5} 
\multirow{3}{*}{\makecell{Comorbidities \\ (2:1)}} & Y & 0.52 & 0.52 & \textbf{0.81} \\ \cline{2-5} 
  & N & 0.56 & 0.54 & \textbf{0.61} \\ \cline{2-5} 
  & Average & 0.54 & 0.53 & \textbf{0.71} \\ \cline{1-5} 
  \multirow{3}{*}{\makecell{Specified HCW \\ (1:6)}} & Y & 0.26 & 0.35 & \textbf{0.44} \\ \cline{2-5} 
  & N & \textbf{0.95} & 0.93 & 0.90 \\ \cline{2-5} 
  & Average & 0.61 & 0.64 & \textbf{0.67} \\ \hline
\end{tabular}
\end{table}

\par{\textbf{What are beneficial parameter settings for automated line listing?}}

To identify which parameter settings are beneficial for automated line listing, we looked at the
best parameter configuration (see Table~\ref{tab:ben_param}) of {\gdlsgns} and {\gdlsghs} which achieved the accuracy values in 
Tables~\ref{tab:qs},~\ref{tab:acc_table} and~\ref{tab:f1}. In Table~\ref{tab:ben_param}, we explored the standard settings of each 
word2vec parameter (dimensionality of word embeddings, window size, negative samples and training iterations) in accordance with previous
research~\cite{levyparam}. Regarding dimensionality of word embeddings, {\gdlsghs} prefers $600$ dimensions, whereas
{\gdlsgns} prefers $300$ dimensions. For the window size, both the models seem to benefit from smaller-sized (5) context windows. 
The number of negative samples is applicable only for {\gdlsgns} where it seems to prefer a single negative
sample. Finally, for the training iterations, both the models benefit from more than 1 training iteration. This is expected as the WHO corpus
used in the template learning process (see section~\ref{sec:expts}) is a smaller-sized corpus with a vocabulary of only $\mathcal{W} = 4447$ 
words. In such scenarios, word2vec models ({\skipnegative} or {\skiph}) generate improved embeddings with higher number 
of training iterations. Finally, both the models are also associated with the parameter $K$ which refers to the
number of indicators $K + 1$ used for extracting the disease onset and clinical features. As expected, the models prefer at least 5 indicators, along with the seed indicator to be used for automated line listing.
Using higher number of indicators increases the chance of discovering an informative indicator for a line list feature.

\begin{table}[!ht]
\centering
\caption{Parameter settings in {\gdlsgns} and {\gdlsghs} for which both the models
achieve optimal performance in terms of average QS and individual feature accuracies
corresponding to MERS line list in Saudi Arabia. Non-applicable combinations are marked by 
\textit{NA}}
\label{tab:ben_param}
\begin{tabular}{|c|c|c|c|c|c|}
\hline
Models & \begin{tabular}[c]{@{}c@{}}Dimensionality \\ (300:600)\end{tabular} & \begin{tabular}[c]{@{}c@{}}Window \\ size \\ (5:10:15)\end{tabular} & \begin{tabular}[c]{@{}c@{}}Negative \\ samples \\ (1:5:15)\end{tabular} & \begin{tabular}[c]{@{}c@{}}Training \\ Iterations\\ (1:2:5)\end{tabular} & \begin{tabular}[c]{@{}c@{}}Indicators \\ ($K$ = 3:5:7)\end{tabular} \\ \hline
\begin{tabular}[c]{@{}c@{}}{\fullmodel}\\  ({\skiph})\end{tabular} & 600 & 5 & \textit{NA} & 5 & 7 \\ \cline{1-6} 
\makecell{{\fullmodel} \\ ({\skipnegative})} & 300 & 5 & 1 & 2 & 5 \\ \hline
\end{tabular}
\end{table}

\par{\textbf{Which indicator keywords discovered using word2vec contribute to the improved 
performance of {\fullmodel}?}}

\begin{figure*}[t!]
  \centering
  \begin{subfigure}{0.33\textwidth}
    \centering
    \includegraphics[width=1.0\linewidth]{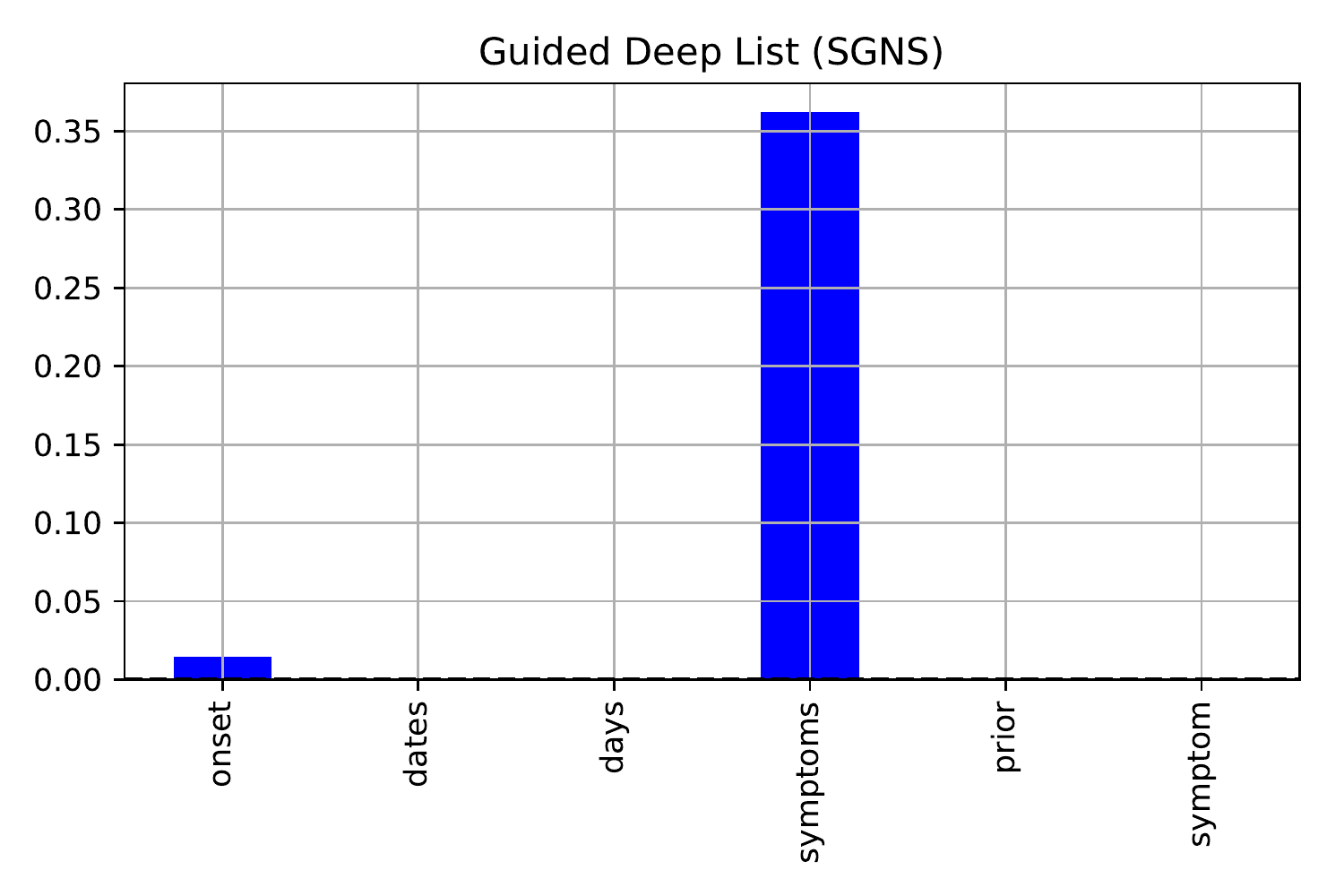}
    \caption{Onset date}
    \label{fig:onset}
  \end{subfigure}
  \begin{subfigure}{0.33\textwidth}
    \centering
    \includegraphics[width=1.0\linewidth]{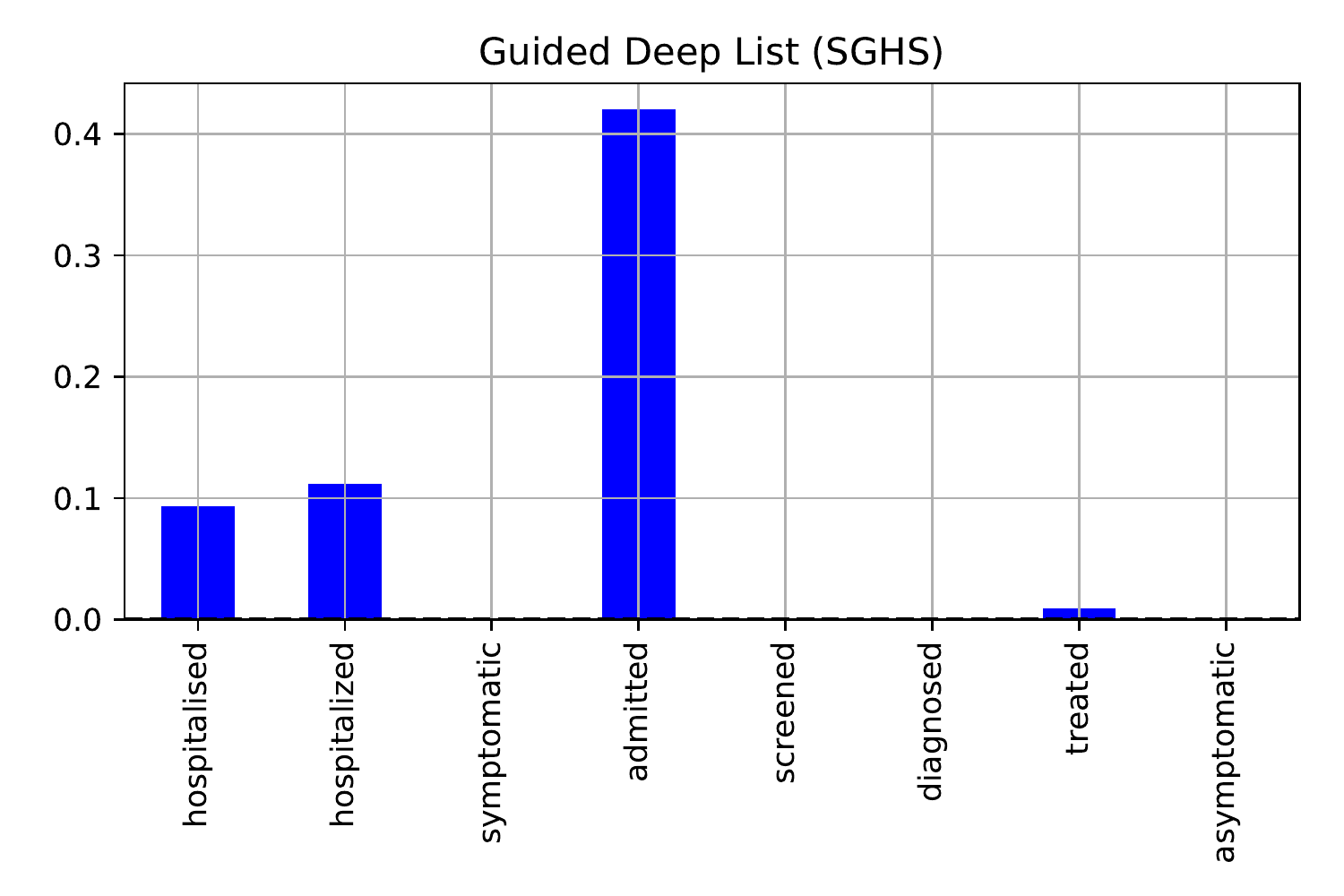}
    \caption{Hospital date}
    \label{fig:hospital}
  \end{subfigure}
  \begin{subfigure}{0.33\textwidth}
    \centering
    \includegraphics[width=1.0\linewidth]{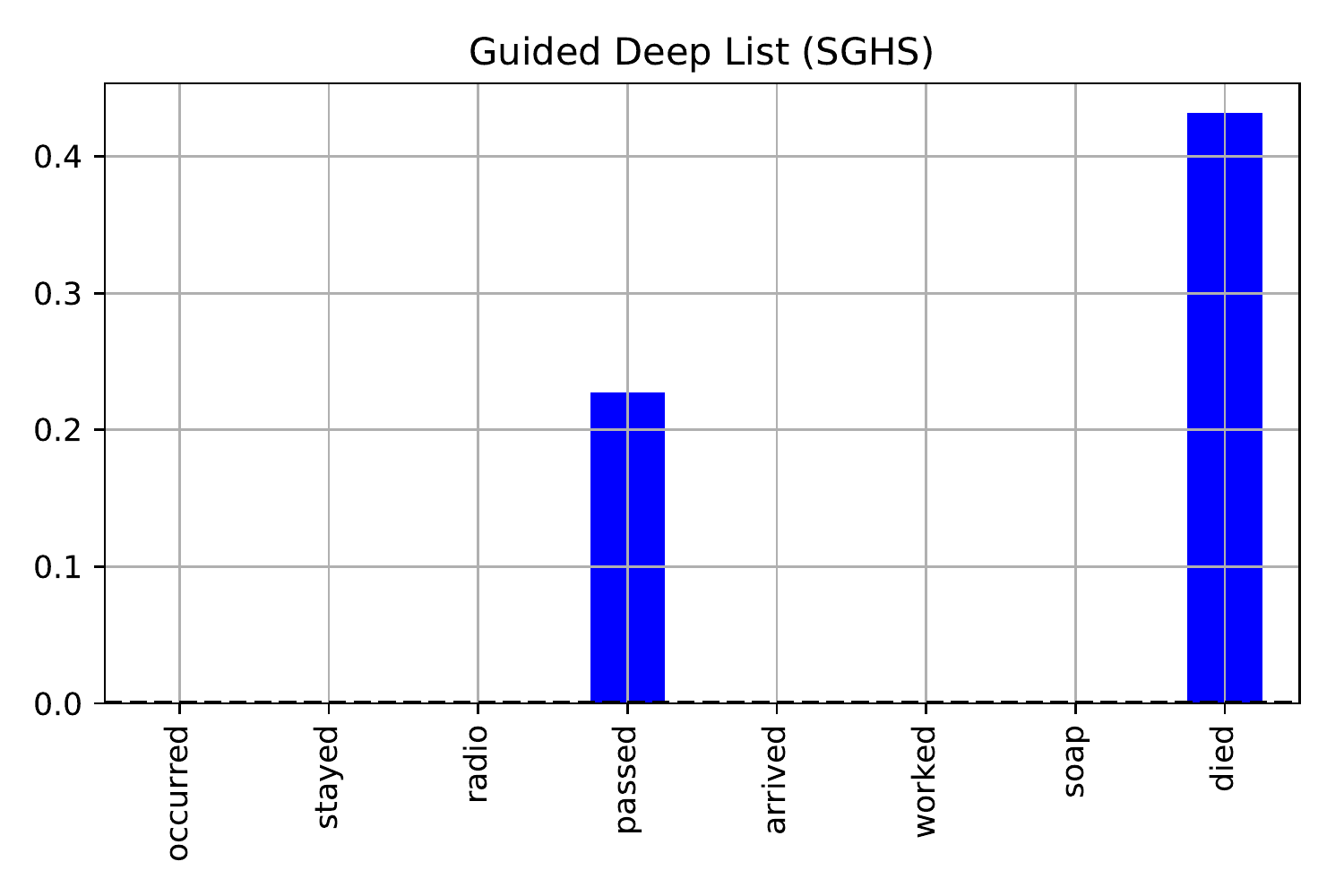}
    \caption{Outcome date}
    \label{fig:outcome}
  \end{subfigure}
  \\
  \begin{subfigure}{0.33\textwidth}
    \centering
    \includegraphics[width=1.0\linewidth]{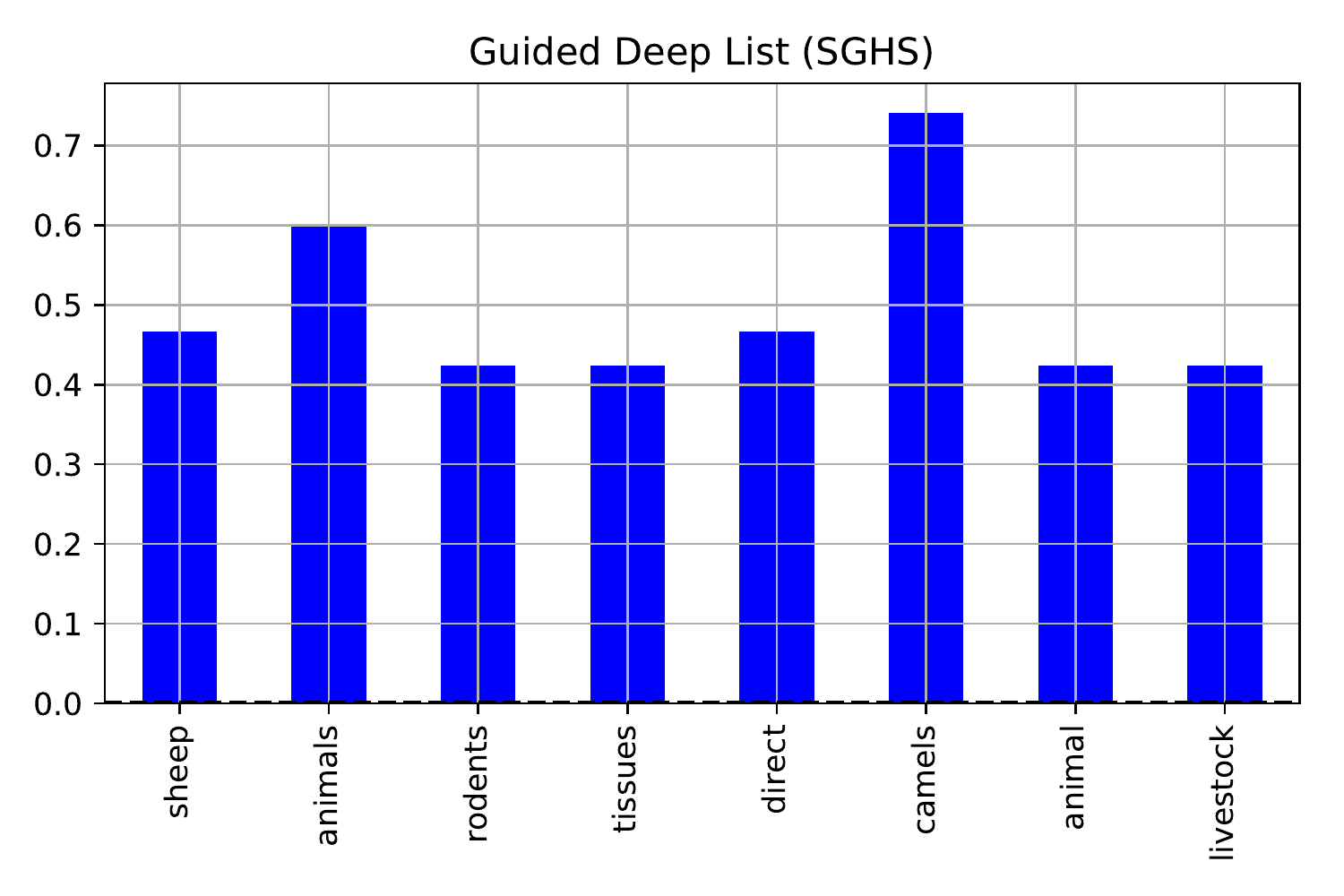}
    \caption{Animal contact}
    \label{fig:animal}
  \end{subfigure}
  \begin{subfigure}{0.33\textwidth}
    \centering
    \includegraphics[width=1.0\linewidth]{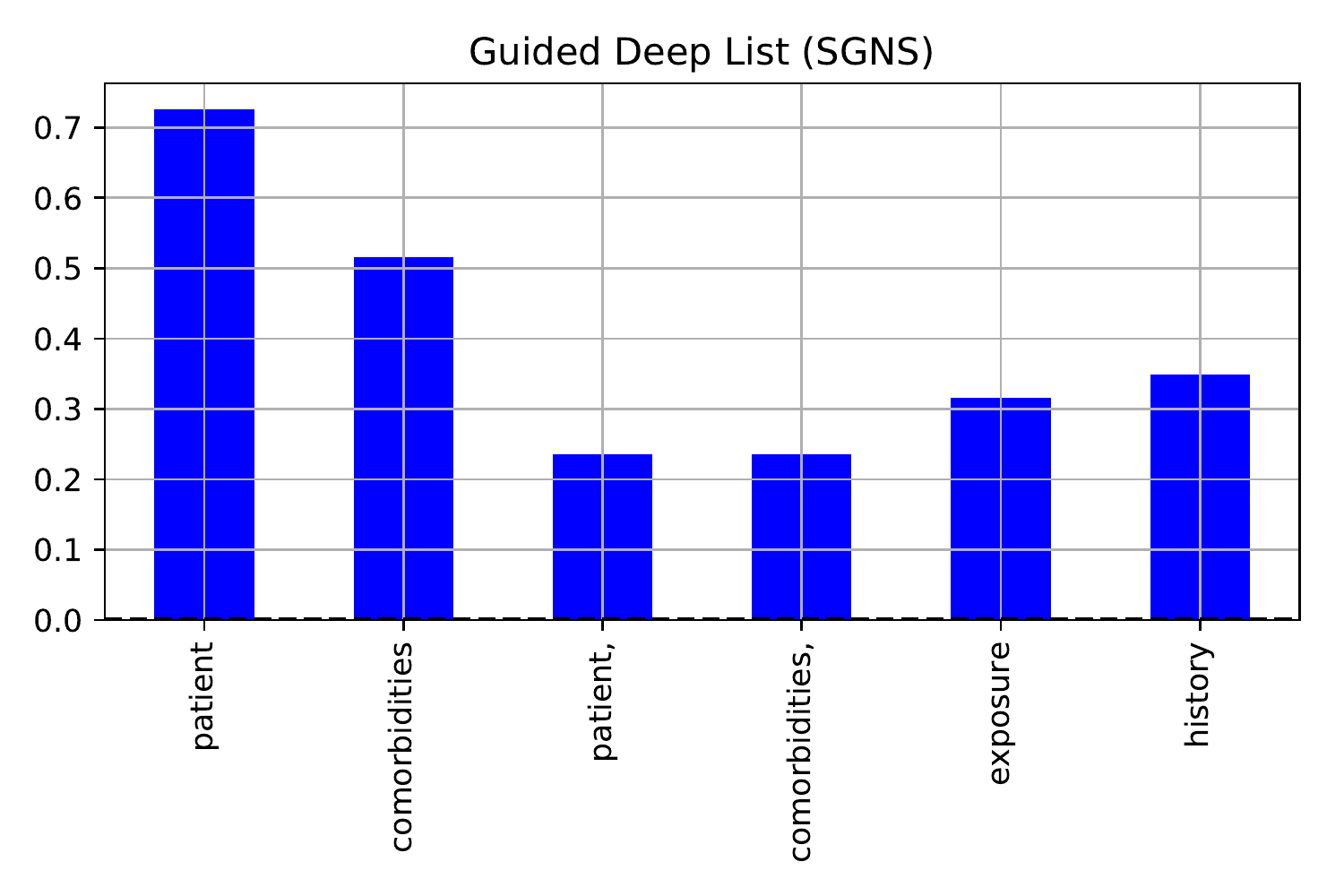}
    \caption{Comorbidities}
    \label{fig:comorbidities}
  \end{subfigure}
  \begin{subfigure}{0.33\textwidth}
    \centering
    \includegraphics[width=1.0\linewidth]{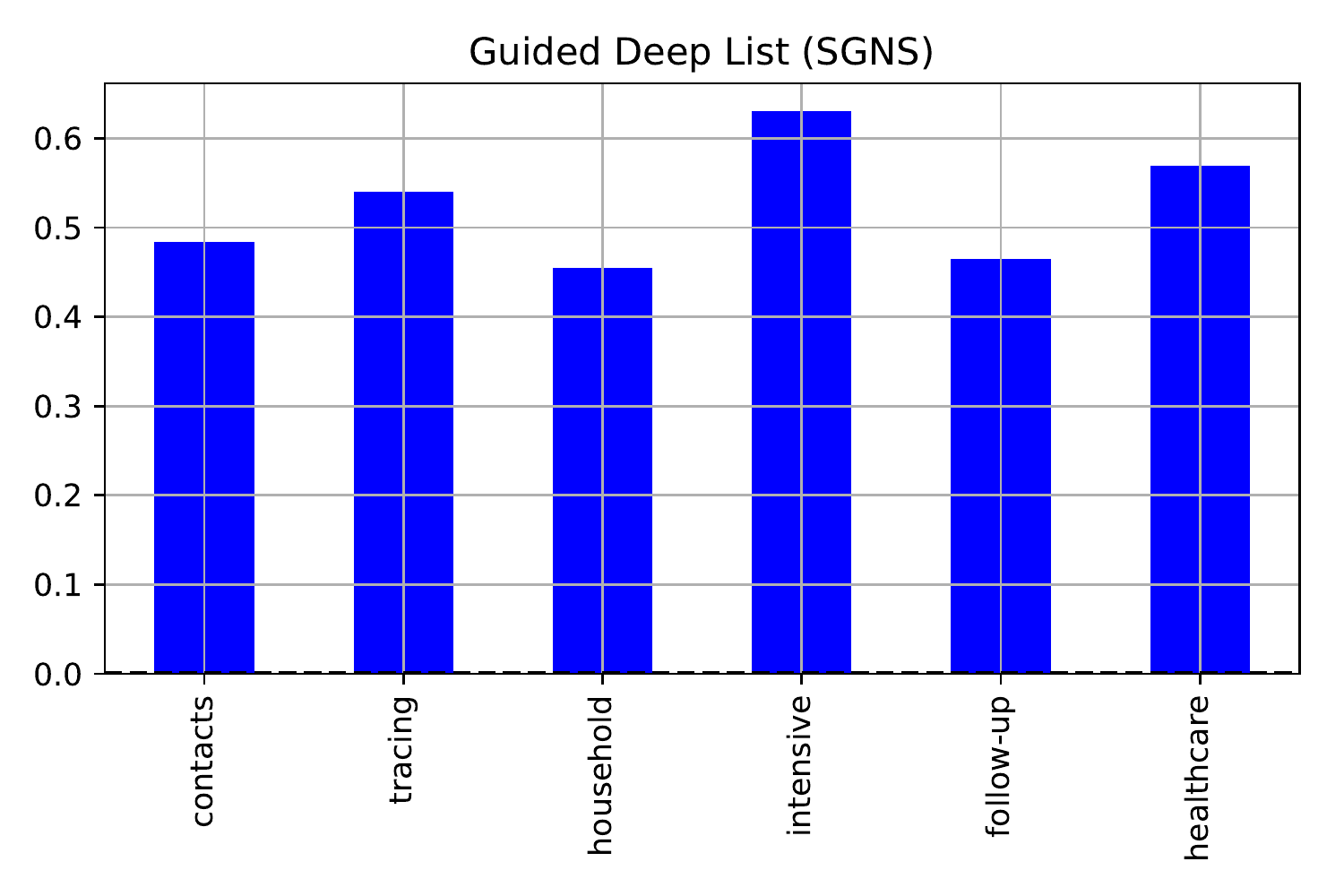}
    \caption{Specified HCW}
    \label{fig:HCW}
  \end{subfigure}
  \caption{Accuracy of individual indicators (including the seed indicator) discovered via word2vec
  methods in {\gdlsgns} or {\gdlsghs} for each line list feature. For clinical features, we show the average F1-score. 
  This figure depicts the informative indicators (indicators showing higher accuracies or F1-scores) which contribute 
  to the improved performance of {\gdlsgns} or {\gdlsghs} for a particular feature. E.g.\ for animal contact, the most 
  informative indicator contributing to the superior performance of {\gdlsghs} is \textit{camels} followed by \textit{animals}
  (seed), \textit{sheep} and \textit{direct}}
  \label{fig:features}
\end{figure*}

Next, we investigate the informative indicators discovered using word2vec which contribute to the improved
performance of {\gdlsgns} or {\gdlsghs} in Tables~\ref{tab:acc_table} and~\ref{tab:f1}. In Figure~\ref{fig:features},
we show the accuracies (or, average F1-score) of individual indicators (including the seed indicator) corresponding 
to the best performing model for a particular line list feature. Regarding onset date (see Figure~\ref{fig:onset}),
{\gdlsgns} is the best performing model and the seed indicator provided as input is \textit{onset}. We observe that
\textit{symptoms} is the most informative indicator achieving an accuracy of 0.36 similar to the overall accuracy 
(see Table~\ref{tab:acc_table}). Rest of the indicators (including the seed indicator) achieve negligible accuracies
and therefore, do not contribute to the overall performance of {\gdlsgns}. Similary, for hospitalization date with 
the seed keyword \textit{hospitalization} provided as input, {\textit{admitted}} emerges out to be most informative 
indicator followed by the seed indicator, \textit{hospitalised} and \textit{treated} (see Figure~\ref{fig:hospital}). 
Finally, for the outcome date, \textit{died} (seed indicator) and \textit{passed} are the two most informative indicators
as observed in Figure~\ref{fig:outcome}.

Regarding the clinical features, we show the average F1-score of individual indicators. For animal contact, the seed indicator provided as input is \textit{animals}. We observe in Figure~\ref{fig:animal}
that the most informative indicator for animal contact is \textit{camels} followed by indicators such as \textit{animals} (seed),
\textit{sheep} and \textit{direct}. This shows that contact with \textit{camels} is the major transmission mechanism for MERS 
disease. The informative
indicators found for comorbidities are \textit{patient}, \textit{comorbidities} and \textit{history}. Finally, regarding specified
HCW, the informative indicators discovered are \textit{healthcare} (seed), \textit{tracing} and \textit{intensive}.

\par{\textbf{Does indirect negation detection play an useful role in extracting clinical features?}}

In level 2 modeling for extracting clinical features, both direct and indirect negation detection are used. For more details, please see section~\ref{sec:methods}. To identify if indirect negation detection contributes positively, we compared the performance of {\fullmodel} with and without indirect negation detection for each clinical feature in Table~\ref{tab:indirect} by reporting the F1-score for each class as well as average F1-score. We observe that indirect negation detection has a positive effect on the performance for animal contact and secondary contact. However, for comorbidities and specified HCW, indirect negation detection plays an insignificant role.

\begin{table}[!ht]
\centering
\caption{Comparing the performance of {\fullmodel} on extraction of clinical features with 
or without indirect negation for MERS line list in Saudi Arabia. It can be seen that
indirect negation improves the performance of {\fullmodel} for animal
contact and secondary contact.}
\label{tab:indirect}
\begin{tabular}{|c|c|c|c|}
  \hline
  Clinical Feature & Class & Direct Negation & Direct + Indirect Negation \\ \hline
  \multirow{3}{*}{Animal contact} & Y & 0.56 & \textbf{0.63} \\ \cline{2-4} 
  & N & 0.80 & \textbf{0.90} \\ \cline{2-4} 
  & Average & 0.68 & \textbf{0.77} \\ \hline
  \multirow{3}{*}{Secondary contact} & Y & \textbf{0.55} & 0.54 \\ \cline{2-4} 
  & N & 0.65 & \textbf{0.72} \\ \cline{2-4} 
  & Average & 0.60 & \textbf{0.63} \\ \hline
  \multirow{3}{*}{Comorbidities} & Y & \textbf{0.86} & 0.82 \\ \cline{2-4} 
  & N & \textbf{0.64} & 0.62 \\ \cline{2-4} 
  & Average & \textbf{0.75} & 0.72 \\ \hline
 \multirow{3}{*}{Specified HCW} & Y & \textbf{0.44} & \textbf{0.44} \\ \cline{2-4} 
 & N & \textbf{0.90} & \textbf{0.90} \\ \cline{2-4} 
 & Average & \textbf{0.67} & \textbf{0.67} \\ \hline
\end{tabular}
\end{table}

\par{\textbf{What insights can epidemiologists gain about the MERS disease from automatically extracted line lists?}}

\begin{figure}[ht!]
  \centering
  \begin{subfigure}{0.5\textwidth}
    \centering
    \includegraphics[width=1.0\linewidth]{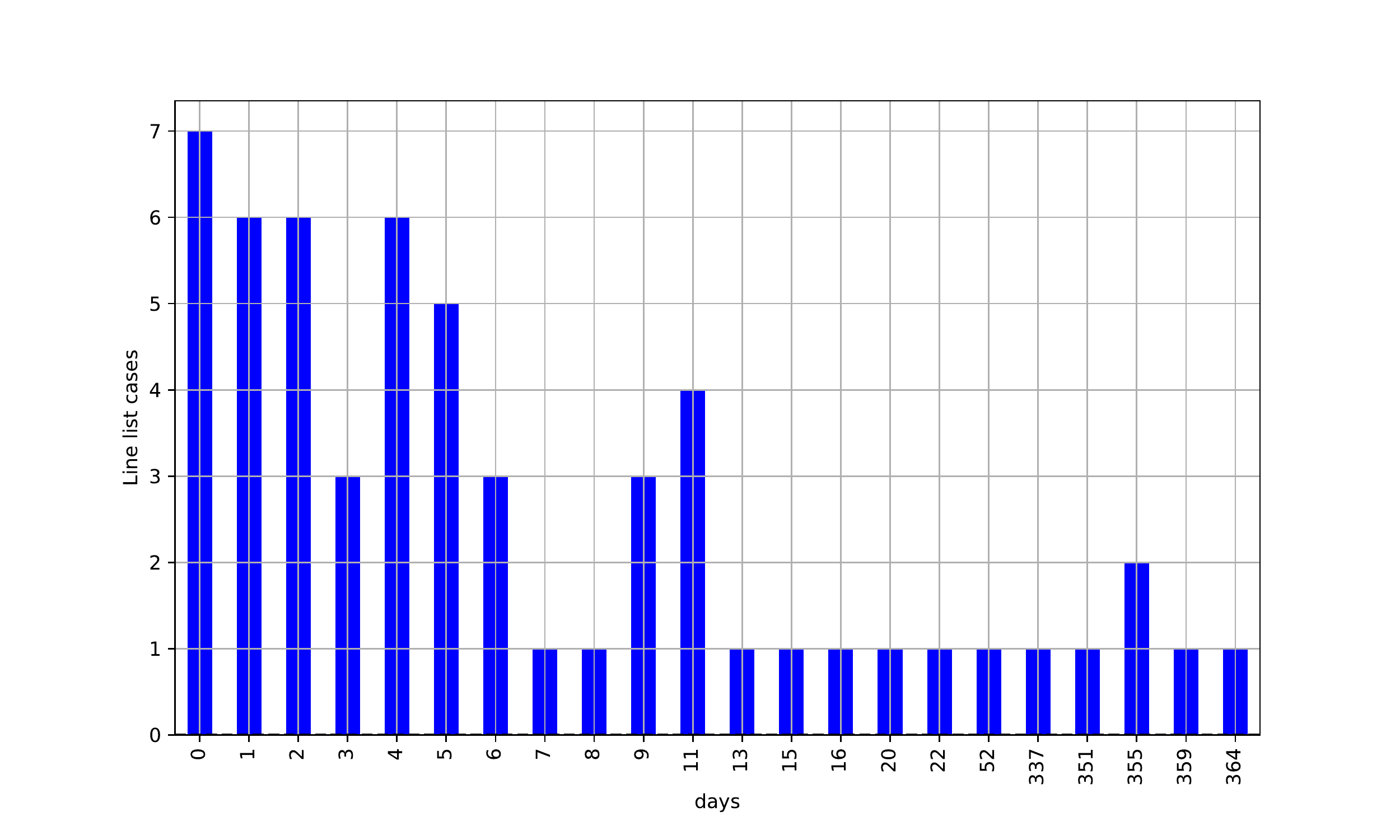}
    \caption{\centering Symptoms-to-hospitalization period distribution}
    \label{fig:inho}
  \end{subfigure}%
  \begin{subfigure}{0.5\textwidth}
    \centering
    \includegraphics[width=1.0\linewidth]{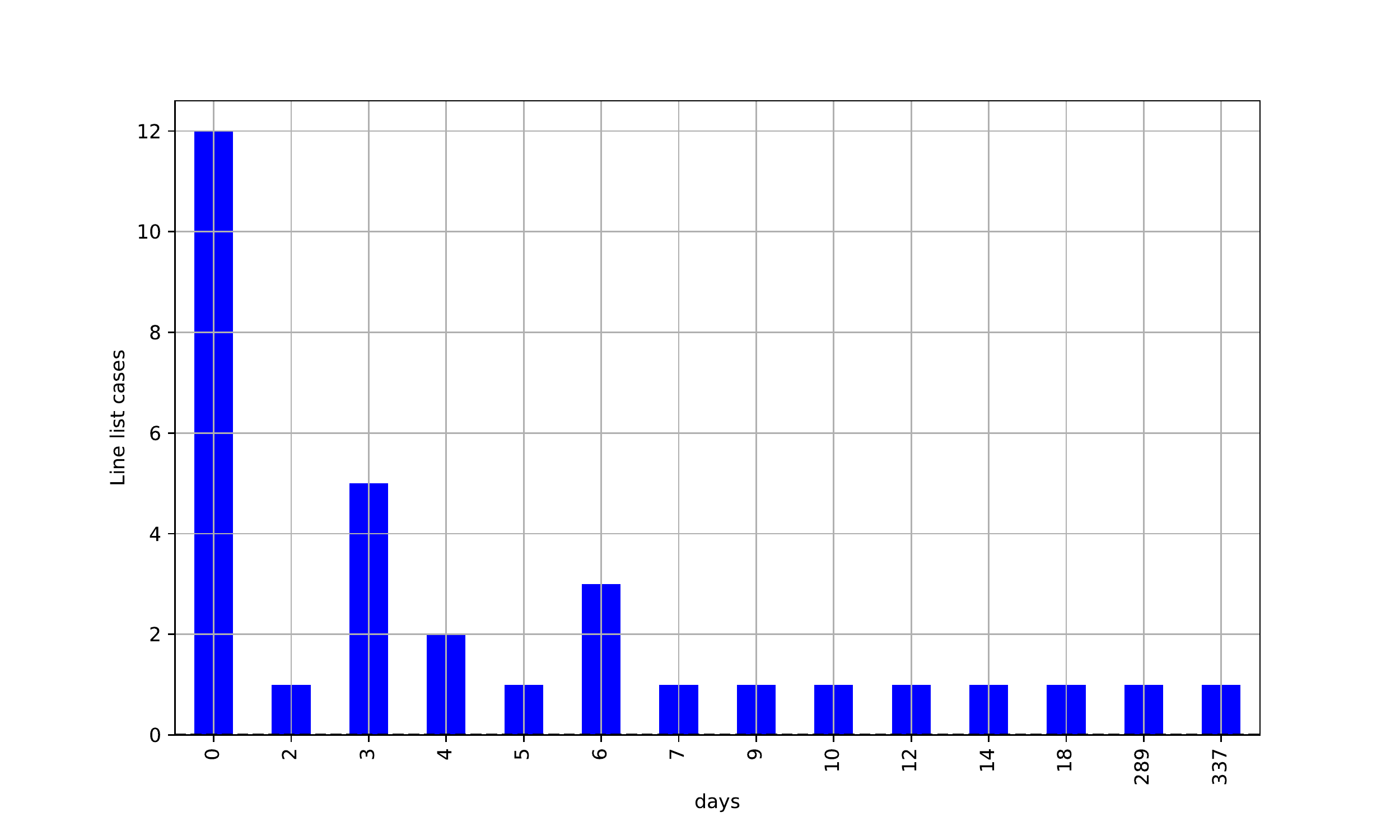}
    \caption{\centering Hospitalization-to-outcome period distribution}
    \label{fig:hoou}
  \end{subfigure}
  \caption{Analysis of disease onset features in the extracted line list}
\end{figure}

Finally, we show some of the utilities of automated line lists by inferring
different epidemiological insights from the line list extracted by {\fullmodel}.
\\
\textbf{Demographic distribution.} In Figure~\ref{fig:ll_overall},
we show the age and gender distribution of the affected individuals in the extracted line list. We observe that males are more prone to getting infected by MERS rather than females. This is expected as males have a higher probability of getting contacted with infected animals (animal contact) or with each other (secondary contact). Also individuals aged between 40 and 70 are more prone to getting infected as evident from the age distribution.
\\
\textbf{Analysis of disease onset features.} We analyzed the symptoms-to-hospitalization period by analyzing the difference (in days) between onset date and hospitalization date in the extracted line list as shown in Figure~\ref{fig:inho}. We observe that most of the affected individuals with onset of symptoms got admitted to the hospital either on the same day or within 5 days. This depicts a prompt responsiveness of the concerned health authorities in Saudi Arabia in terms of admitting the individuals showing symptoms of MERS\@. In Figure~\ref{fig:hoou}, we also show a distribution of the hospitalization-to-outcome period (in days). Interestingly, we see that the distribution has a peak at 0 which indicates that most of the infected individuals admitted to the hospital died on the same day indicating high fatality rate of MERS case.


\section*{Acknowledgements}
{Supported by the Intelligence Advanced Research Projects Activity
  (IARPA) via Department of Interior National Business Center (DoI/NBC)
  contract number D12PC000337, the US Government is authorized to
  reproduce and distribute reprints of this work for Governmental
  purposes notwithstanding any copyright annotation thereon.
  Disclaimer: The views and conclusions contained herein are those of
  the authors and should not be interpreted as necessarily representing
  the official policies or endorsements, either expressed or implied, of
  IARPA, DoI/NBC, or the US Government.
}

\section*{Supplementary Information}

  Codes and data for this manuscript are available at \url{https://github.com/sauravcsvt/KDD_linelisting}.


\end{document}